\documentclass{article}

\usepackage[final]{neurips_2024}

\usepackage[acronym, nonumberlist, toc]{glossaries}
\usepackage[utf8]{inputenc}
\usepackage[T1]{fontenc}
\usepackage{hyperref}
\usepackage{url}
\usepackage{booktabs}
\usepackage{amsfonts}
\usepackage{nicefrac}
\usepackage{microtype}
\usepackage{xcolor}
\usepackage{amsmath}
\usepackage{graphicx}
\usepackage{multirow}
\usepackage{verbatim}

\newacronym{crossentropy}{CEL}{cross-entropy loss}
\newacronym{fft}{FFT}{full fine-tuning}
\newacronym{glue}{GLUE}{}
\newacronym{kd}{KD}{knowledge distillation}
\newacronym{kl}{KL}{Kullback-Leibler divergence}
\newacronym{kdlora}{KD-LoRA}{}
\newacronym{llm}{LLM}{large language model}
\newacronym{lora}{LoRA}{low-rank adaptation}
\newacronym{peft}{PEFT}{parameter-efficient fine-tuning}
\makeglossaries

\title{\gls{kdlora}: A Hybrid Approach to Efficient
Fine-Tuning with LoRA and Knowledge Distillation}

\author{
  Rambod Azimi\textsuperscript{1} \\
  \And
  Rishav Rishav\textsuperscript{1,}\textsuperscript{3} \\
  \And
  Marek Teichmann\textsuperscript{2}
  \And
  Samira Ebrahimi Kahou\textsuperscript{1,}\textsuperscript{3,}\textsuperscript{4} \\
}

\begin{document}
\maketitle

\renewcommand{\thefootnote}{\arabic{footnote}}
\footnotetext[1]{Mila – Quebec AI Institute, \textsuperscript{2}{CM Labs Simulations Inc, \textsuperscript{3}{University of Calgary, \textsuperscript{4}{Canada CIFAR AI Chair}}}}

\begin{abstract}
\Glspl{llm} have demonstrated remarkable performance across various downstream tasks. However, the high computational and memory requirements of \glspl{llm} are a major bottleneck. To address this, \gls{peft} methods such as \gls{lora} have been proposed to reduce computational costs while ensuring minimal loss in performance. Additionally, \gls{kd} has been a popular choice for obtaining compact student models from teacher models. In this work, we present \gls{kdlora}, a novel fine-tuning method that combines \gls{lora} with \gls{kd}. Our results demonstrate that \gls{kdlora} achieves performance comparable to \gls{fft} and \gls{lora} while significantly reducing resource requirements. Specifically, \gls{kdlora} retains 98\% of \gls{lora}'s performance on the \gls{glue} benchmark, while being 40\% more compact. Additionally, \gls{kdlora} reduces GPU memory usage by 30\% compared to \gls{lora}, while decreasing inference time by 30\% compared to both \gls{fft} and \gls{lora}. We evaluate KD-LoRA across three encoder-only models: BERT, RoBERTa, and DeBERTaV3. Code is available at \href{https://github.com/rambodazimi/KD-LoRA}{https://github.com/rambodazimi/KD-LoRA}.
\end{abstract}

\section{Introduction}
\label{introduction}
With advancements in transformer~\citep{vaswaniattention} architectures and hardware capabilities, including GPUs and distributed training, researchers have been able to develop \glspl{llm} with billions of parameters~\citep{li2020pytorchdistributedexperiencesaccelerating, narayanan2021efficientlargescalelanguagemodel, dash2023optimizingdistributedtrainingfrontier}, such as LLaMA 3.1~\citep{dubey2024llama3} which boasts up to 405 billion parameters. These models, trained on trillions of tokens, exhibit remarkable capabilities across various downstream tasks~\citep{brown2020languagemodelsfewshotlearners, liu2019robertarobustlyoptimizedbert, wei2022finetunedlanguagemodelszeroshot}. However, fine-tuning these models requires substantial energy and memory demands~\citep{samsi2023wordswattsbenchmarkingenergy}. Furthermore, in recent years, the growth in the number of parameters in \glspl{llm} has significantly outpaced the advancements in available GPU memory~\citep{lialin2023scalingscaleupguide}, amplifying the challenges of managing memory during fine-tuning~\citep{singh2024studyoptimizationsfinetuninglarge, kim2024llmemestimatinggpumemory, dong2024finetuningdeployinglargelanguage}.

\begin{figure}[ht]
    \centering
    \includegraphics[width=1\linewidth]{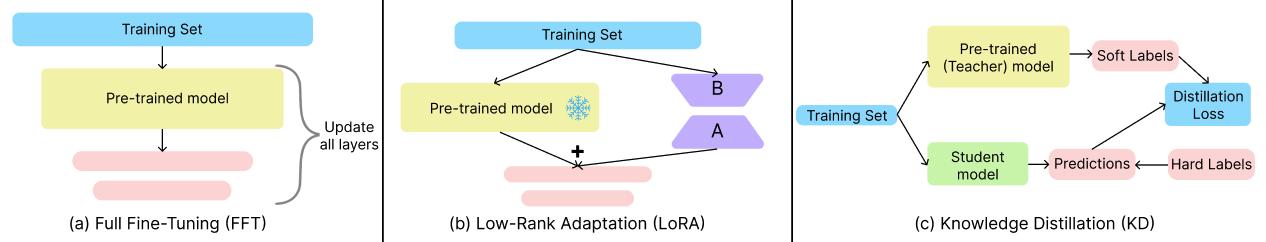}
    \caption{Overview of three fine-tuning methods: (a) \Gls{fft}, which updates all model parameters; (b) \gls{lora}, which adds low-rank matrices to update a small subset of parameters; and (c) \gls{kd}, which trains a smaller student model to emulate a larger teacher model.}
    \label{fig:Fine-Tuning Techniques}
\end{figure}

To address these challenges, \gls{peft} techniques~\citep{houlsby2019parameterefficienttransferlearningnlp} have emerged as effective solutions, which fine-tune a small subset of parameters while keeping the majority fixed. As shown in Figure~\ref{fig:Fine-Tuning Techniques}, one popular \gls{peft} technique, \gls{lora}~\citep{hu2021loralowrankadaptationlarge}, and its variants~\citep{zhang2023adaloraadaptivebudgetallocation, zi2023deltalorafinetuninghighrankparameters, ren2024analyzingreducingcatastrophicforgetting, zhao2024continualforgettingpretrainedvision, liu2024doraweightdecomposedlowrankadaptation} reduce the number of trainable parameters by introducing small, trainable rank decomposition matrices, maintaining performance as \gls{fft} across many tasks~\citep{dettmers2023qloraefficientfinetuningquantized}.
For example, DoRA~\citep{liu2024doraweightdecomposedlowrankadaptation} enhances LoRA by decomposing pre-trained weights into magnitude and direction, applying LoRA to directional updates for reduced trainable parameters. Similarly, AdaLoRA~\citep{zhang2023adaloraadaptivebudgetallocation} improves LoRA by dynamically allocating parameters based on their importance, optimizing efficiency and performance, particularly under tight budget constraints.

However, the effectiveness of \gls{peft} methods varies across \glspl{llm} based on several factors such as model architecture and task type~\citep{pu2023empiricalanalysisstrengthsweaknesses, lee2023surgicalfinetuningimprovesadaptation}. Additionally, \gls{lora} still requires substantial memory, as it does not reduce the activation memory cost compared to \gls{fft}~\citep{chen2016trainingdeepnetssublinear, zhang2023lorafamemoryefficientlowrankadaptation, zhao2024galorememoryefficientllmtraining}. For example, a GPT-like model with 1.5 billion parameters, a sequence length of 1K, and a batch size of 32 requires approximately 60 GB of GPU memory~\citep{rajbhandari2020zeromemoryoptimizationstraining}. Moreover, \gls{lora} does not improve inference time, as the full model still needs to be processed during inference~\citep{liao2023parameterefficientfinetuningintroducingnew, gu2024sarasingularvaluebasedadaptive}.

\Gls{kd}~\citep{hinton2015distillingknowledgeneuralnetwork} has become another prominent way to make the training and inference less memory-intensive by transferring capabilities of larger teacher models, such as GPT-4~\citep{openai2024gpt4technicalreport}, Gemini~\citep{geminiteam2024geminifamilyhighlycapable}, and LLaMA~\citep{dubey2024llama3} to smaller student models without greatly compromising performance~\citep{gu2024minillmknowledgedistillationlarge, xu2024surveyknowledgedistillationlarge}.

\Gls{kd} has for instance been used to distill the BERT model into TinyBERT~\citep{jiao2020tinybertdistillingbertnatural} that has only 14.5 million parameters without significant performance loss.
The performance of the distilled 11B parameter T5 model~\citep{hsieh2023distillingstepbystepoutperforminglarger} has been shown to even surpass that of the much larger 540B parameter PaLM teacher model.

In this paper, we introduce \gls{kdlora}, a novel fine-tuning method that integrates \gls{lora} into the \gls{kd} framework to achieve competitive performance with reduced computational costs, making it ideal for deployment in resource-limited environments. We accomplish this by incorporating \gls{lora} matrices into the student model and then applying the distillation process while updating the \gls{lora} matrices of the student model. By combining \gls{kd} with \gls{lora}, we leverage the strengths of both methods: \gls{lora}’s efficiency in reducing trainable parameters and \gls{kd}’s ability to effectively transfer knowledge to more compact student models, resulting in reduced model size and shorter inference time.

We evaluate the effectiveness of \gls{kdlora} in comparison to \gls{fft} and \gls{lora} across three encoder-only \glspl{llm}: BERT \citep{devlin2019bertpretrainingdeepbidirectional}, RoBERTa \citep{liu2019robertarobustlyoptimizedbert}, and DeBERTaV3 \citep{he2023debertav3improvingdebertausing}. For the \gls{kd} component, we select a smaller student model from the same family for each \gls{llm}. For each \gls{glue} benchmark task \citep{wang2019gluemultitaskbenchmarkanalysis}, we explore various hyperparameter configurations and \gls{peft} settings, utilizing NVIDIA A100 GPUs. The median performance is reported based on the top 6 configurations. Our comprehensive experiments on the \gls{glue} benchmark reveal that \gls{kdlora} offers several advantages:

\begin{itemize}
    \item \gls{kdlora} achieves about 97\% of \gls{fft}’s performance while updating significantly fewer parameters. For instance, \gls{fft} fine-tunes all 110M parameters of the BERT-base model, whereas \gls{kdlora} fine-tunes only 1.2M parameters with a rank of 8.
    \item \gls{kdlora} achieves about 98\% of \gls{lora}’s performance while incorporating knowledge from a larger teacher model and using fewer trainable parameters due to the more compact student model. For example, \gls{lora} fine-tunes 2.9M parameters of the RoBERTa-base model with a rank of 8, whereas \gls{kdlora} fine-tunes only 1.5M parameters with the same rank.
    \item \gls{kdlora} is 40\% more compact than both \gls{fft} and \gls{lora} by utilizing a smaller student model. This approach also reduces GPU memory usage by approximately 75\% compared to \gls{fft} and 30\% compared to \gls{lora} during training.
    \item \gls{kdlora} reduces inference time by approximately 30\% while maintaining the same convergence speed as both \gls{fft} and \gls{lora}.
\end{itemize}

\section{Method}
\label{method}
We propose \gls{kdlora}, a novel fine-tuning methodology that integrates \gls{lora} with \gls{kd}. The proposed method involves three main steps: (1) selecting and fine-tuning a teacher model, (2) initializing a smaller student model with \gls{lora} modules, and (3) performing distillation to transfer knowledge from the teacher model to the student model.

\textbf{Teacher Model}. Let \(\mathcal{T}\) denote the teacher model, initialized from a pre-trained language model (e.g., BERT, RoBERTa, DeBERTa). The teacher model is fine-tuned on a specific task \(\mathcal{D}_{\text{task}}\), using \gls{fft}, where all parameters of the model are updated~\citep{lv2024parameterfinetuninglargelanguage}. The objective function for fine-tuning the teacher model is:

\begin{equation}
\mathcal{L}_{\text{task}}^{\mathcal{T}} = \frac{1}{|\mathcal{D}_{\text{task}}|} \sum_{(x_i, y_i) \in \mathcal{D}_{\text{task}}} \mathcal{L}_{\text{CE}}(\mathcal{T}(x_i), y_i)
\end{equation}

where \(\mathcal{L}_{\text{CE}}\) denotes the \gls{crossentropy} loss, \(x_i\) represents the input data, and \(y_i\) denotes the corresponding label. This loss function measures the discrepancy between the predicted probabilities \(\mathcal{T}(x_i)\) and the true labels \(y_i\). The fine-tuned teacher model \(\mathcal{T}\) then serves as the source of distilled knowledge.

\textbf{Student Model with \gls{lora}}. The student model \(\mathcal{S}\) is initialized from a smaller version within the same model family as the teacher model \(\mathcal{T}\). We modify the student model by injecting \gls{lora} modules into its architecture. Specifically, \gls{lora} is applied to the attention layers, where the weight matrices \(W_q\) and \(W_v\) (corresponding to the query and value projections) are decomposed as follows:

\begin{equation}
W_q = W_q^{\text{base}} + A_q B_q, \quad W_v = W_v^{\text{base}} + A_v B_v
\end{equation}

where \(W_q^{\text{base}}\) and \(W_v^{\text{base}}\) are the pre-trained weight matrices, while \(A_q\), \(B_q\), \(A_v\), and \(B_v\) are the low-rank matrices, the only parameters updated during fine-tuning.

\textbf{\gls{kdlora}}. With \gls{lora} modules already in place, the \gls{kd} process is performed, where the student model \(\mathcal{S}\) learns from the teacher model. During this phase, the student model, equipped with \gls{lora}, adapts its low-rank matrices to capture the knowledge transferred from the teacher. The student model is trained on the target task \(\mathcal{D}_{\text{task}}\) using the combined loss function \(\mathcal{L}_{\text{total}}^{\mathcal{S}}\), which is given by:

\begin{equation}
\mathcal{L}_{\text{total}}^{\mathcal{S}} = \alpha \mathcal{L}_{\text{task}}^{\mathcal{S}} + (1 - \alpha) \mathcal{L}_{\text{\gls{kd}}}(z^{\mathcal{S}}, z^{\mathcal{T}})
\label{eq:total_loss}
\end{equation}

where \(z^{\mathcal{T}}\) and \(z^{\mathcal{S}}\) are the logits (outputs before the softmax layer) of the teacher and student models, respectively. The \gls{kd} loss \(\mathcal{L}_{\text{\gls{kd}}}\) is computed as the \gls{kl}~\citep{shlens2014noteskullbackleiblerdivergencelikelihood} between the softened output probabilities of the teacher model \(\mathcal{T}\) and the student model \(\mathcal{S}\). The parameter \(\alpha\) balances the task-specific loss \(\mathcal{L}_{\text{task}}^{\mathcal{S}}\) and the \gls{kd} loss \(\mathcal{L}_{\text{\gls{kd}}}\). During each training step, the student model’s low-rank matrices are updated to minimize the loss in Eq.~\ref{eq:total_loss}.

\section{Experiments}
\label{experiments}

\begin{table}[ht]
  \caption{
Performance metrics for BERT-base (BERT-b), DeBERTa-v3-base (DeB-b), and RoBERTa-base (RoB-b) across \gls{glue} tasks using three fine-tuning methods. Results show the median of the top 6 hyperparameter and \gls{peft} setups. DistilBERT-base (DBERT-b), DeBERTa-v3-small (DeB-s), and DistilRoBERTa-base (DRoB-b) serve as student models. Metrics include Matthews correlation for CoLA, average Pearson/Spearman correlations for STS-B, average accuracy/F1 scores for MRPC and QQP, and accuracy for all other tasks. \textbf{\gls{kdlora} achieves about 97\% of \gls{fft}'s performance and about 98\% of \gls{lora}'s performance.}}
  \label{llm-performance}
  \centering
  \resizebox{\textwidth}{!}{
    \begin{tabular}{lccccccccccc}
      \toprule
      \textbf{Model} & \multicolumn{2}{c}{\textbf{BERT-b}} & \multicolumn{1}{c}{\textbf{DBERT-b}} & \multicolumn{2}{c}{\textbf{DeB-b}} & \multicolumn{1}{c}{\textbf{DeB-s}} & \multicolumn{2}{c}{\textbf{RoB-b}} & \multicolumn{1}{c}{\textbf{DRoB-b}}\\
      \cmidrule(lr){2-4} \cmidrule(lr){5-7} \cmidrule(lr){8-10}
      \textbf{Method} & \gls{fft} & \gls{lora} & \gls{kdlora} & \gls{fft} & \gls{lora} & \gls{kdlora} & \gls{fft} & \gls{lora} & \gls{kdlora} \\
      \midrule
      CoLA & 57.7 & 56.9 & 56.3 & 67.8 & 69.1 & 68.1 & 60.9 & 59.4 & 56.8 \\
      \( \text{MNLI}_{m} \) & 84.5 & 83.4 & 82.0 & 90.3 & 90.3 & 88.8 & 87.7 & 87.2 & 83.3 \\
      \( \text{MNLI}_{mm} \) & 84.9 & 83.9 & 82.4 & 90.6 & 90.2 & 89.0 & 87.4 & 86.9 & 83.4 \\
      MRPC & 89.0 & 89.2 & 88.3 & 91.9 & 90.9 & 90.7 & 91.1 & 89.9 & 89.3 \\
      QNLI & 91.8 & 91.1 & 89.7 & 94.1 & 94.3 & 93.4 & 92.7 & 92.8 & 90.7 \\
      QQP & 89.7 & 87.9 & 89.1 & 91.2 & 90.4 & 89.9 & 89.8 & 88.6 & 87.3 \\
      RTE & 71.6 & 70.1 & 64.0 & 85.0 & 84.0 & 78.8 & 74.8 & 71.8 & 65.3 \\
      SST-2 & 92.8 & 92.6 & 92.0 & 95.9 & 96.0 & 95.7 & 94.3 & 94.2 & 92.9 \\
      STS-B & 89.5 & 88.9 & 88.7 & 91.5 & 91.1 & 89.8 & 90.8 & 90.3 & 87.9 \\
      WNLI & 56.3 & 56.9 & 56.3 & 66.9 & 56.3 & 56.3 & 56.3 & 56.3 & 56.3 \\
      \midrule
      \textbf{Score} & \textbf{80.8} & \textbf{80.1} & \textbf{78.9} & \textbf{86.5} & \textbf{85.3} & \textbf{84.1} & \textbf{82.6} & \textbf{81.7} & \textbf{79.3} \\
      \bottomrule
    \end{tabular}
  }
\end{table}

\begin{table}[ht]
\centering
\caption{Comparison of trainable parameters, memory usage, and inference time for three fine-tuning methods across three models and their distilled counterparts for \gls{kdlora}. Inference time is averaged over 100 runs on the CoLA validation set. \textbf{With a rank of 8, \gls{kdlora} reduces trainable parameters by 99\% compared to \gls{fft} and 49\% compared to \gls{lora}, while lowering GPU memory usage by 75\% and 30\%, respectively. \gls{kdlora} also cuts inference time by 30\%.}}
\resizebox{\textwidth}{!}{
    \begin{tabular}{lcccccccc}
    \toprule
    \textbf{Model} & \textbf{Method} & \textbf{Rank 8} & \textbf{Rank 16} & \textbf{Rank 32} & \textbf{Rank 64} & \textbf{Memory Usage} & \textbf{Inference Time} \\
    \midrule
    BERT-base & \gls{fft} & 110M & 110M & 110M & 110M & 1332.0MB & 6.10s \\
    & \gls{lora} & 2.9M & 5.9M & 11.8M & 23.6M & 463.5MB & 6.22s \\
    DistilBERT-base & \gls{kdlora} & 1.2M & 2.4M & 4.7M & 9.4M & \textbf{296.8MB} & \textbf{5.36s} \\
    \midrule
    RoBERTa-base & \gls{fft} & 125M & 125M & 125M & 125M & 1515.9MB & 7.21s \\
    & \gls{lora} & 2.9M & 5.9M & 11.8M & 23.6M & 531.9MB & 7.19s \\
    DistilRoBERTa-base & \gls{kdlora} & 1.5M & 2.9M & 5.9M & 11.8M & \textbf{358.3MB} & \textbf{4.44s} \\
    \midrule
    DeBERTa-v3-base & \gls{fft} & 183M & 183M & 183M & 183M & 2234.5MB & 14.37s \\
    & \gls{lora} & 2.9M & 5.9M & 11.8M & 23.6M & 763.4MB & 15.62s \\
    DeBERTa-v3-small & \gls{kdlora} & 1.5M & 2.9M & 5.9M & 11.8M & \textbf{590.3MB} & \textbf{10.38s} \\
    \bottomrule
    \end{tabular}
}
\label{training_parameters}
\end{table}

\begin{figure}[ht]
    \centering
    \includegraphics[width=1\linewidth]{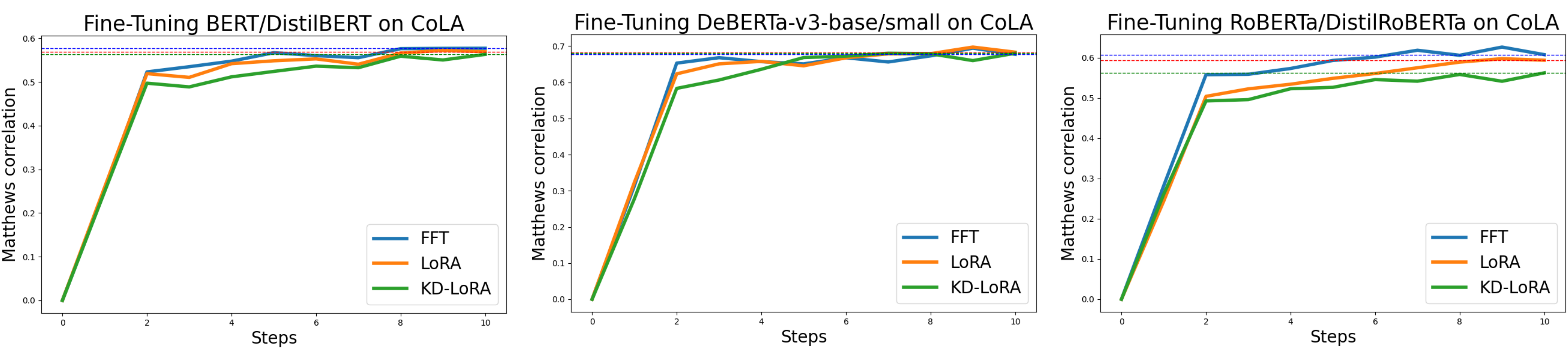}
    \caption{Comparison of convergence speed between full fine-tuning (\gls{fft}), \gls{lora}, and \gls{kdlora} for three \glspl{llm} on the CoLA task. \textbf{\gls{kdlora} matches the convergence speed of \gls{fft} and \gls{lora}.}}
    \label{fig:Fine-Tuning Result}
\end{figure}

For our experiments, we select three widely recognized encoder-only \glspl{llm}: BERT~\citep{devlin2019bertpretrainingdeepbidirectional}, RoBERTa~\citep{liu2019robertarobustlyoptimizedbert}, and DeBERTaV3~\citep{he2023debertav3improvingdebertausing}. We evaluate three fine-tuning strategies across these models on the \gls{glue} benchmark: \gls{fft}, \gls{lora}, and \gls{kdlora}. In this approach, we employ compact student models that belong to the same family as their corresponding larger teacher models. Specifically, we use DistilBERT-base~\citep{sanh2020distilbertdistilledversionbert}, DeBERTa-v3-small, and DistilRoBERTa-base as student models for BERT-base, DeBERTa-v3-base, and RoBERTa-base, respectively. For \gls{fft}, we select 25 hyperparameter configurations, varying learning rates (2e-5 to 5e-5), batch sizes (8 to 32), epochs (2 to 5), and weight decay (0.01 to 0.1). For \gls{lora} and \gls{kdlora}, we select 24 \gls{peft} configurations, varying rank (8 to 32), epochs (3 to 5), \gls{lora} alpha (16 to 32), and \gls{lora} dropout (0.0 to 0.1). All experiments are conducted using NVIDIA A100 GPUs. Table~\ref{llm-performance} shows the results calculated based on the median of the top 6 configurations. Table~\ref{training_parameters} provides the number of trainable parameters for each method at different ranks, along with their GPU memory usage during inference and the inference time calculated on the CoLA dataset.

\Gls{kdlora} achieves approximately 97\% of \gls{fft}’s performance and about 98\% of \gls{lora}’s, with scores of 78.9 for the student model of BERT-base compared to 80.8 for \gls{fft} and 80.1 for \gls{lora}. It reduces the number of trainable parameters by about 99\% compared to \gls{fft} and about 49\% compared to \gls{lora}, updating 1.5M parameters in the DistilRoBERTa-base model with \gls{kdlora} versus 2.9M with \gls{lora} at a rank of 8. \gls{kdlora} also reduces GPU memory usage by 75\% compared to \gls{fft} and 30\% compared to \gls{lora}, resulting in a model that is about 40\% more compact than both \gls{fft} and \gls{lora}. Additionally, \gls{kdlora} decreases inference time by around 30\% on the CoLA dataset, while maintaining comparable convergence speed, as illustrated in Figure~\ref{fig:Fine-Tuning Result}.

\section{Conclusion}
\label{conclusion}
We present \gls{kdlora}, a novel fine-tuning method that integrates \gls{lora} modules into a student model while leveraging \gls{kd} from a larger teacher model. Empirical results on the \gls{glue} benchmark show that \gls{kdlora} retains approximately 97\% of \gls{fft} performance and 98\% of \gls{lora} performance, all while reducing model size by around 40\%. \gls{kdlora} also lowers trainable parameters by 99\% compared to \gls{fft} and 49\% compared to \gls{lora}, reduces GPU memory usage by 75\% compared to \gls{fft} and 30\% compared to \gls{lora}, and cuts inference time by 30\%.

\subsubsection*{Acknowledgements}
The authors thank CMLabs, Mila and CIFAR for research funding. This research was enabled by the compute provided by Calcul Quebec and Digital Research Alliance of Canada.

%\section*{References}
%\bibliographystyle{unsrtnat}%
\bibliography{references}

\end{document}